\documentclass[letterpaper]{article} 
\usepackage[]{aaai2026}  
\usepackage{times}  
\usepackage{helvet}  
\usepackage{courier}  
\usepackage[hyphens]{url}  
\usepackage{graphicx} 
\usepackage{natbib}  
\usepackage{caption} 
\usepackage{algorithm}
\usepackage{algorithmic}

\usepackage{newfloat}
\usepackage{listings}
\DeclareCaptionStyle{ruled}{labelfont=normalfont,labelsep=colon,strut=off} 
\floatstyle{ruled}
\newfloat{listing}{tb}{lst}{}
\floatname{listing}{Listing}
\usepackage{booktabs}       
\usepackage{amsfonts}       
\usepackage{nicefrac}       
\usepackage{microtype}      
\usepackage{xcolor}         

\usepackage{enumerate}
\usepackage{multirow}

\title{Planning in the LLM Era: Building for Reliability and Efficiency}

\author {
Michael Katz,  
Harsha Kokel,
Kavitha Srinivas, 
Shirin Sohrabi
}

\affiliations{
IBM Research
}

\newcommand{\inlinecite}[1]{\citeauthor{#1}~\shortcite{#1}}

\begin{document}

\maketitle

\begin{abstract}
Growing attention to intelligent agents has put a spotlight on one of their central capabilities: planning. Early attempts to leverage large language models (LLMs) for planning relied on single-shot plan generation, followed by hybrid approaches that coupled LLMs with limited external search. These methods, unsound and incomplete by their very nature, often require substantial resources without yielding better solutions on unseen problems. As the limitations of LLMs become clearer, recent work has shifted toward using them at \emph{solution construction time} -- generating symbolic solvers for a family of problems that can be verified and then used efficiently at inference time.  This trend reflects the growing need for agents that are both reliable and resource-efficient. It also offers a path towards generating maintainable planners with minimal dependence on language models at inference time. In this paper, we argue that this shift reflects a broader realignment of the planning field in the LLM era. We examine three major categories of planner-generation methods, discuss their current limitations, and outline research steps towards a more reliable and efficient LLM-based generation of planners. 
\end{abstract}

\section{Introduction}

Planning is one of the defining capabilities of intelligent agents, enabling them to reason about actions, anticipate their consequences, and synthesize solutions that achieve complex goals. This recognition has recently brought renewed attention to planning within the broader surge of interest in language-driven agents. Early attempts to use large language models (LLMs) for planning treated each task independently by prompting the model to generate entire plans in a single step \cite{silver-et-al-neurips2022-wsfmdm,valmeekam-et-al-neurips2023datasets,valmeekam-et-al-neurips2023, kambhampati-et-al-icml2024}. These approaches achieved limited success on simple problems but quickly revealed fundamental limitations: the inability to reconsider earlier decisions, poor long-horizon reasoning, and difficulty generalizing to unseen instances. More recent evaluations of frontier models reflect the rapid and multifaceted nature of progress in LLM-based planning, showing that systems such as GPT-5 can approach the coverage of a strong planner like LAMA on existing domains, while still degrading under obfuscation and requiring vastly greater computational resources \cite{correa2025frontierLLMs}. Hybrid methods that incorporated restricted forms of backtracking or external search into LLM-based exploration \cite{yao-et-al-neurips2023,Besta2023GraphOT,Algo_of_thoughts} are unsound and incomplete, and often require dramatically more LLM calls without delivering reliably better performance \cite{katz-et-al-neurips2024}. Further, all that hard work is wasted -- each problem is solved separately and no computation is reused for the next problem in the same dataset.

\begin{figure}[t]
  \centering
  \includegraphics[width=0.4\textwidth]{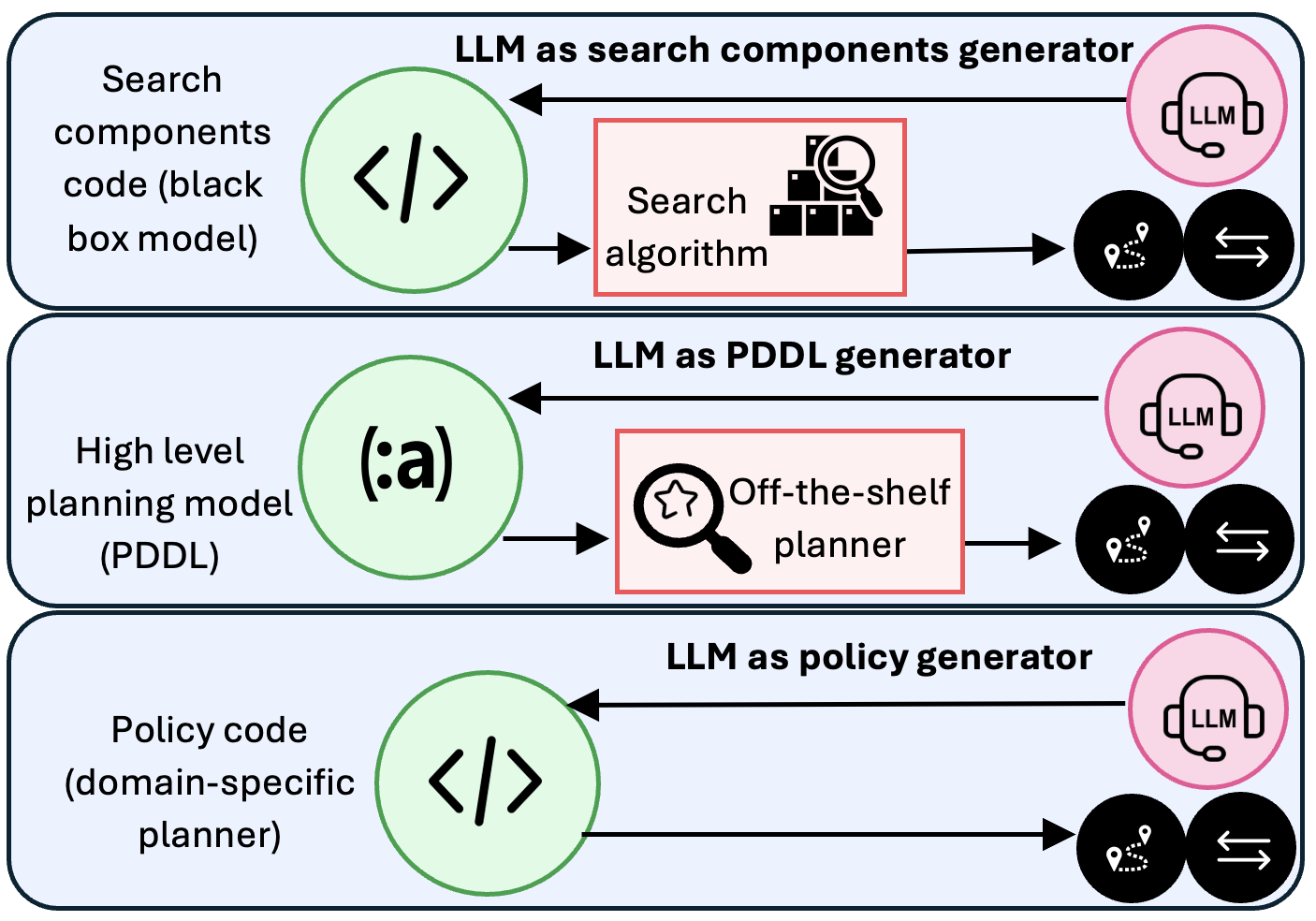}
  \caption{
  An overview of the planner generation methods
  }
  \label{fig:approaches}
\end{figure}

These realizations, coupled with increased awareness of the environmental and monetary costs of LLM inference gave rise to the emergence of a promising paradigm: instead of relying on LLMs at inference time, use them at construction time. In this view, LLMs are not planners themselves but generators of domain-specific planners that can be validated, maintained, and run efficiently without repeated model calls.  This shift is visible across multiple lines of work. In this position paper, we discuss three directions outlined in Figure \ref{fig:approaches}. Search-based planner generation (NL2Search) uses LLMs to generate code for search components, i.e., successor generator and goal test \cite{katz-et-al-neurips2024,cao-et-al-neurips2024wsowa}, as well as heuristic functions \cite{tuisov-et-al-arxiv2025,correa-et-al-neurips2025}. Planning via planning models generation (NL2PDDL) focuses on translating natural language task description into formal planning language that can be consumed by off-the-shelf planners \cite{guan-et-al-neurips2023,oswald-et-al-icaps2024,gestrin-et-al-arxiv2024,huang-et-al-aaai2025,tantakoun-et-al-acl-findings2025}. Policy code generation (NL2Policy) leverages LLMs to synthesize domain-specific strategies and executable policies that generalize across problem instances \cite{silver-et-al-aaai2024,hodel-bsc2024,stein-et-al-arxiv2025}. We discuss the current state of these directions and the next steps that must be addressed to operationalize these ideas in trustworthy and effective language-driven agents.

\section{Generating Search Based Planners}

An emerging trend in solving planning tasks with language models focused on simply asking the model to produce a plan by providing a few examples \cite{silver-et-al-neurips2022-wsfmdm}. These models were able to find plans only in the simplest cases, when the solutions were very similar in structure to the example plans. Any need to generate something novel or unseen resulted in failure \cite{valmeekam-et-al-neurips2023}.
This quickly reinforced a familiar insight from classical planning: generating a plan often requires search. Hence, an alternative line of work shifted toward integrating search, the ability to backtrack and consider many possibilities before finding a valid one. As language models excel at text-based input, the focus was on planning problems expressed in natural language. A variety of approaches were proposed, exploiting search methods for the heavy lifting of exploring the problem’s state space, while the language model was used to define the search space, with the required search components successor function $succ$ and goal test $is\_goal$ implemented as calls to the model \cite{hao-et-al-emnlp2023,Reflexion_ShinnCGNY23,yao-et-al-neurips2023,Besta2023GraphOT,Algo_of_thoughts}.

All these attempts have the same common pitfall, their search components implementation calls a language model, which takes significant time and is impractical in planning, where problems have combinatorially large state spaces and it is imperative to have an {\em efficient} implementation of these search components.
The aforementioned literature made a different choice, instead heavily restricting the search algorithms, bounding their exploration depth, width, frontier size, etc. Consequently, the algorithms lost their {\em soundness} and {\em completeness} properties \cite{katz-et-al-neurips2024}, without gaining in precision \cite{katz-et-al-arxiv2025}. A more promising alternative has been to use LLMs only at construction time to generate code (e.g., Python code) that implements search components, avoiding LLM calls during search or inference time.
Such implementations can then be tested and reflected upon, improving gradually with case-specific feedback, like in the case of (Automated) Thought of Search (ToS/AutoToS) \cite{katz-et-al-neurips2024,cao-et-al-neurips2024wsowa}.

That initial work on using the language models to obtain code for the $succ$ and $is\_goal$ search components has paved the way for the next step, obtaining code for a domain-specific heuristic $h$ \cite{correa-et-al-neurips2025,tuisov-et-al-arxiv2025}. That was a major step, allowing to move away from a blind exploration into the realm of an {\em informed} search, making the generated solvers substantially more efficient. Both of these investigations focus on deterministic tasks expressed in PDDL, one for purely propositional and the other for numeric planning. In both cases, it should be rather straightforward to move away from PDDL to tasks that are specified purely in natural language.

\subsection{The Next Steps}
While progress on generating search-based planner components is encouraging, these methods are not yet ready for realistic settings. The foundation laid by ToS and AutoToS demonstrates that code for search components can be synthesized and iteratively refined, but several aspects of these approaches remain in their simplest form. One major example is the {\bf state representation}. In ToS, the state representation was hand-crafted and somewhat similar to a multi-valued encoding \cite{backstrom-nebel-compint1995}, while in heuristic-generation work it was propositional and derived from a formal task specification. This raises the question of whether certain representations work better for generating search components. More importantly, it raises the question of where state features should come from in the first place. Prior work assumes that state features are known, and relaxing this assumption is a critical step toward adapting to new domains, such as webpage navigation in WebArena \cite{ZhouX0ZLSCOBF0N24} and API calling in AppWorld \cite{TrivediKHMDLGSB24}. 
One might expect that language models could automatically infer such state features, but current models struggle with this task \cite{vafa-et-al-arxiv2024}. Nonetheless, prior work on learning state representations from images \cite{asai-fukunaga-aaai2018}, text \cite{lindsay-et-al-icaps2017,sohrabi-et-al-aaai2018}, and reinforcement learning \cite{konidaris2018skills,EchchahedC25} offers potential clues for addressing this challenge.
A central question here is how to determine the right {\bf level of abstraction} for both the actions and the state representation.

Other limitations arise when dealing with {\bf partial information}. Generating successors might require interaction with the environment, like in the aforementioned AppWorld. Such cases might be handled either via planning and execution loop or via knowledge compilation \cite{palacios-geffner-jair2009}. Regardless, it can be challenging to determine the initial state, goal formulation, or even whether additional information-gathering actions are required.

Another limitation of the implementation strategy used in AutoToS is its {\bf linear} nature, where each single alternative suggested by the language model is accepted without comparison. In more complex domains, generating and evaluating multiple candidate components should be a global search process with the ability to backtrack on earlier decisions.

\section{Planning via PDDL Generation}

PDDL \cite{mcdermott-et-al-tr1998} has long served as de-facto the standard formalism for expressing planning tasks, providing the input for a wide range of planners. However, representing a planning task in PDDL is one of the few remaining challenges that require manual human labor. Recent research has turned to language models for help with closing this gap by translating natural language task descriptions into formal PDDL specifications. The developed approaches, that we collectively refer to as NL2PDDL, aim to make planning more accessible, while reducing the burden of manual knowledge engineering and enabling the use of existing planners in settings where formal domain specifications in PDDL were not yet readily available. The plan generation therefore can be done with existing domain-independent planners, exploiting decades of research and engineering efforts in making efficient planners.

Among the first is the work of \inlinecite{guan-et-al-neurips2023}, which explores using pre-trained LLMs to construct PDDL (domain and problem). The approach takes as input the detailed natural language description of the task, in-context examples, description of the domain, any physical constraints, and a dynamically updated list of predicates that is initially empty. 
The LLM then incrementally proposes action argument lists, preconditions, and effects, together with any newly introduced predicates and their natural language descriptions, and refines them through validator feedback and human guidance. To reduce the reliance on human experts for the in-the-loop feedback, \inlinecite{huang-et-al-aaai2025} propose a fully automated method. They would first generate a diverse library of candidate action schemas to capture multiple interpretations of natural language task description, then apply semantic filtering using sentence encoders to automatically validate, filter, and rank the generated action schemas. Their experiments show that this approach can produce sound plans and better accommodate natural language ambiguity without human feedback. These approaches laid the foundation for the technical construction of the feedback flows. The work of \inlinecite{oswald-et-al-icaps2024} focused on evaluation of constructed models, going beyond the semantic filtering and addressing syntactic validity, semantic correctness, and usability of the generated results by existing planners. It did so by focusing on the operational semantics instead of action schema similarity, comparing models in terms of similarity of the solution spaces they implicitly encode.
Another approach by \inlinecite{gestrin-et-al-arxiv2024} aims for robustness from minimal textual specifications (rather than assuming rich textual specifications). Their framework, NL2Plan, incrementally extracts necessary information to build the PDDL planning task in a particular order, similar to how students are taught to model in PDDL.

To learn more about the plethora of available techniques that use language models to help construct planning models, one can look at the survey by  \inlinecite{tantakoun-et-al-acl-findings2025}. They also introduce Language-to-Plan (L2P), an open-source Python library that re-implements several NL2PDDL approaches. L2P is designed to streamline reuse and enable systematic, reproducible experimental results across a variety of NL2PDDL approaches.

\subsection{The Next Steps}
Since this method deals with generating models in a restrictive form, all the shortcomings of the search-based methods described in the previous section persist here as well. Choosing the right level of abstraction for both actions and predicates/numeric fluents as well as how to deal with features outside of the classical fragment (full observability, deterministic action dynamics, and offline access to task structure) are arguably the biggest open questions. The interplay between planning and execution often provides information beyond the common knowledge - strong suit of language models - that might be exploited via means of transformation. To give an example, in AppWorld, determining whether a concrete {\em Venmo} payment action is applicable may require observing the response of an API call, which cannot be known a priori.
Another limitation is that some applications require object creation \cite{correa-et-al-icaps2024}, conditional plans and loops, some of which might need to be handled outside of the planning problem, for example in AppWorld, sending {\em Venmo} payments to “all” friends, formulating an appropriate goal state requires quantifying over an unknown number of objects, a well-known limitation of classical PDDL.
Furthermore, it is not clear how to integrate the procedural glue code that is required between APIs, or actions in this case, at certain abstraction levels, such as dealing with multi-page API responses, transforming returned data formats, or chaining heterogeneous tool outputs. All of these fall outside the capabilities of current NL2PDDL pipelines.

Beyond addressing these modeling limitations, there is also substantial room for improving the feedback used during PDDL construction. Existing approaches typically explore modifications to PDDL models in a greedy, linear fashion, evaluating each proposed model in isolation rather than with respect to the model it aims to refine. This prevents systems from determining whether a newly proposed model represents a true improvement. While some work begins to move beyond single-path refinement \cite{huang-et-al-aaai2025}, a more systematic model-space search—such as the one explored in \cite{caglar-et-al-aaai2024} is still largely missing.

There is also considerable room to improve how semantic errors are detected and corrected. Recent work on knowledge engineering in the LLM era \cite{vallati2025keps} emphasizes that knowledge engineering is not solely about drafting action schemas but involves iterative refinement, validation, maintenance, and ensuring that models accurately reflect operational semantics. Hybrid workflows, in which LLMs assist human designers and are complemented by principled model-validation and repair techniques, may therefore offer a more robust foundation. Most NL2PDDL systems rely on VAL for syntactic validation, but deeper semantic inconsistencies—incorrect invariants, missing preconditions, unintended side effects—are rarely addressed. Incorporating richer feedback mechanisms, including domain-analysis tools, invariants, or landmarks, could substantially strengthen model quality. Moreover, current pipelines accept a single LLM suggestion at each refinement step; exploring multiple candidate models or applying meta-reasoning over the refinement trajectory could prevent error accumulation and improve convergence.

\section{Planning via Policy Code Generation}

Generalized planning deals with finding policies that can efficiently solve the entire family of planning problems \cite{jimenez-et-al-ker2019}. The conceptual idea of exploring the use of LLMs for generating a generalized policy in the form of code  \cite{silver-et-al-aaai2024} is naturally appealing, as the generated code can be validated, adapted, and when deemed working, deployed. In this original two-phase formulation, a language model is asked to summarize and describe in natural language both the problem and a strategy to solve the problem. Then, a language model is asked to implement the strategy in code. The generated code can be tested with a held out set of instances to produce a feedback in case of an error. A key limitation of such an approach is the inherent assumption that the produced natural language strategy is correct; indeed, while the method works remarkably well on some domains, there are domains where the produced strategy is incorrect \cite{hodel-bsc2024}.  To mitigate this limitation, the follow up work proposed generating the strategy in the form of pseudo-code \cite{stein-et-al-arxiv2025}. The pseudo-code can then be verified with held out instances, asking the language model to emulate an execution of the pseudo-code on these inputs. In this way, mistakes can be detected and corrected prior to the generation of the generalized plan, substantially improving the reliability across diverse IPC domains.

Approaches that leverage LLMs' programming abilities to generate a policy are also emerging in related areas of task and motion planning~\cite{progprompt,AdaPlanner,CodeAspolic} as well as web interaction domains~\cite{song2025coact}. Code-As-Policies~\cite{CodeAspolic} define a hierarchical code generation approach with LLMs such that new functions are defined recursively to create complex policies for robot. Policies ProgPrompt~\cite{WangCY0L0J24} proposes to prompt the language model to implement a python function as a policy for a new task, given a policy as in-context example. CodeActAgent~\cite{WangCY0L0J24} proposes to reduce LLM calls by prompting LLM to provide executable python code that can iterate over multiple actions in a single generation. AdaPlanner~\cite{AdaPlanner} proposed adaptive closed-loop planning with LLMs through code-prompting for AlfWorld, with a refinement flow that is triggered when the generated program fails. In web-interaction domain CoAct~\cite{song2025coact} proposes to query LLM for actions as well as CodeActions, to improve efficiency.

\subsection{The Next Steps}

While the concept of two-phase iterative generation of code policies using language models was established by the original work \cite{silver-et-al-aaai2024}, the adaptation of the first phase to take a form that can be evaluated is significant \cite{stein-et-al-arxiv2025}. Coming up with the strategy and implementing the strategy are the two conceptual steps that need to be performed, but the actual realization can differ. The current work validates the two steps separately, moving to the next once the first one was deemed valid. One can however imagine that during the second step, an evidence is obtained that the strategy/pseudo-code obtained in the first step is incorrect, resulting in revisiting the first step.

Similarly to the case of AutoToS, a limitation of the current approach is its linear nature, where alternatives are weighted against each other and local decisions are made. This local search through the space of alternative changes proposed by the language model is an improvement over the paradigm adopted by AutoToS, but here as well, the better choice would be a global, systematic search procedure.

A major limitation is the restriction to PDDL in the input. While extending to natural language specified inputs would bring forward the many of the issues pointed out in the previous sections, it might be unavoidable in real-world applications.

\section{Cross-Category Observations}

We have discussed each of the categories in isolation. Here, we propose a cross-category view.

 Policy-code generation appears to get us closer to realistic agentic environments, with many of the limitations observed in other approaches being easier to tackle within this paradigm. Unlike NL2Search or NL2PDDL, policy-code generation can directly express procedural glue code, interleave sensing with acting (addressing partial observability more naturally), and incorporate reusable components that arise frequently in real applications. Such components natively capture the hierarchical nature of some applications.
 In domains like AppWorld, for example, one can imagine writing policies that capture sub-tasks, such as sending Venmo payments to friends, co-workers, or roommates.
 It is also possible to define macro-actions or reusable code fragments that support multiple tasks, including authentication, retrieval, filtering, or error handling. 

It is also worth noting that these methods can be used to improve each other. As a simple example, one can imagine NL2Search first used to obtain the search components, passing these generated components code as an additional input to NL2PDDL. More complex iterations are also possible and sometimes are unavoidable.  
In domains where search is needed and pure policy-code generation is not possible, a mix of approaches might work best. Higher-level actions/sub-tasks might be realized via policy-code, while the search-requiring planning can be done via one of the other two methods. This kind of integration of the discussed methods seems to be most promising for handling realistic scenarios.

To sum up, while the progress to date in LLM-driven policy synthesis is promising, opportunities remain for the planning community to shape how planning concepts, policy generation, procedural knowledge, and hierarchical structure, can make meaningful impact in real-world agentic systems.

\section{Conclusions}

In this position paper, we examined three
major categories of planner-generation methods: 
search-based planner generation (NL2Search), planning
via PDDL or model generation (NL2PDDL), and planning via
policy or solver code generation (NL2Policy). The methods
focus on using LLMs at construction time, to generate the search components, the planning models, or policy that can be validated and
used efficiently at inference time. 
We examined each direction, highlighting the progress made to date while outlining the challenges, and the future steps needed to make these approaches usable in real-world applications. 
Although each line of work currently has limitations and/or makes simplifying assumptions, these challenges also reveal opportunities for future research and can help motivate research to address the identified gaps. 
Ultimately, the opportunity ahead is to both advance planning research along these directions and to have a critical view and/or framework for assessing their impact in the LLM era, ensuring the planning contributions meaningfully influence the efficiency, reliability, and capabilities of agentic systems in the years to come.

\fontsize{10pt}{11pt}\selectfont

\end{document}